\documentclass{article}

\usepackage{microtype}
\usepackage{graphicx}
\usepackage{subcaption}
\usepackage{booktabs} 
\usepackage{multirow}
\usepackage{hyperref}
\usepackage{array}
\usepackage{amsmath}
\usepackage{amssymb}
\usepackage{mathtools}
\usepackage{amsthm}
\usepackage{threeparttable,enumitem}
\usepackage{pifont,makecell}

\usepackage[preprint]{icml2026}

\usepackage[capitalize,noabbrev]{cleveref}

\theoremstyle{plain}

\theoremstyle{definition}

\theoremstyle{remark}

\usepackage[textsize=tiny]{todonotes}

\begin{document}

\twocolumn[
  \icmltitle{ELIQ: A Label-Free Framework for Quality Assessment of Evolving AI-Generated Images}



  \icmlsetsymbol{equal}{*}

  \begin{icmlauthorlist}
    \icmlauthor{Xinyue Li}{sjtu}
    \icmlauthor{Zhiming Xu}{xian}
    \icmlauthor{Min Tang}{sjtu}
    \icmlauthor{Zhaolin Cai}{sjtu}
    \icmlauthor{Sijing Wu}{sjtu}
    \icmlauthor{Xiongkuo Min}{sjtu}
    \icmlauthor{Yitong Chen}{sjtu}
    \icmlauthor{Guangtao Zhai}{sjtu}
  \end{icmlauthorlist}

  \icmlaffiliation{sjtu}{Shanghai Jiao Tong University}
  \icmlaffiliation{xian}{Xi'an Jiaotong University}
  
  \icmlcorrespondingauthor{Yitong Chen}{yitongchen@sjtu.edu.cn}
  \icmlcorrespondingauthor{Guangtao Zhai}{zhaiguangtao@sjtu.edu.cn}

  \icmlkeywords{Machine Learning, ICML}

  \vskip 0.3in
]



\printAffiliationsAndNotice{}  

\begin{abstract}

Generative text-to-image models are advancing at an unprecedented pace, continuously shifting the perceptual quality ceiling and rendering previously collected labels unreliable for newer generations.
To address this, we present \textbf{ELIQ}, a \textit{\textbf{\underline{L}}abel-free Framework for \textbf{\underline{Q}}uality Assessment of \textbf{\underline{E}}volving AI-generated \textbf{\underline{I}}mages}.
Specifically, ELIQ focuses on visual quality and prompt-image alignment, automatically constructs positive and aspect-specific negative pairs to cover both conventional distortions and AIGC-specific distortion modes, enabling transferable supervision without human annotations. 
Building on these pairs, ELIQ adapts a pre-trained multimodal model into a quality-aware critic via instruction tuning and predicts two-dimensional quality using lightweight gated fusion and a Quality Query Transformer. 
Experiments across multiple benchmarks demonstrate that ELIQ consistently outperforms existing label-free methods, generalizes from AI-generated content (AIGC) to user-generated content (UGC) scenarios without modification, and paves the way for scalable and label-free quality assessment under continuously evolving generative models. The code will be released upon publication.

\end{abstract}

\section{Introduction}
\label{sec_introduction}

\begin{figure}
    \centering
    \includegraphics[width=0.9\linewidth]{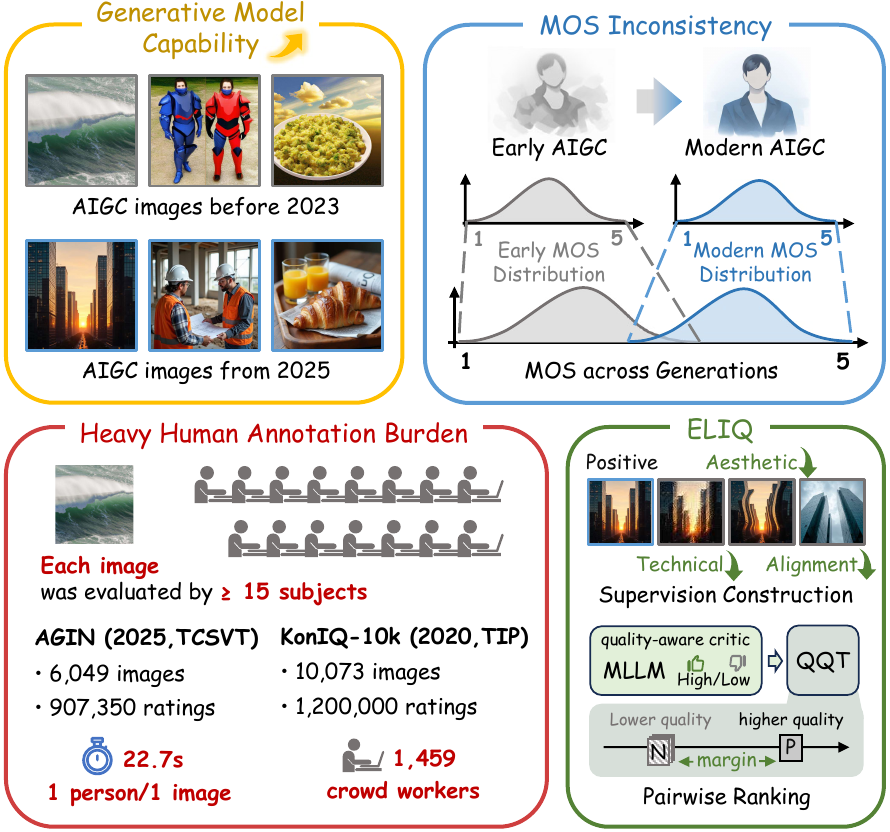}
    \caption{The rapid evolution of generative models shifts MOS distributions, making annotations increasingly inconsistent. ELIQ replaces absolute MOS labels with automatically constructed supervision, enabling scalable quality assessment for evolving AIGC.}
    \label{fig_1}
\end{figure}

The rapid evolution of generative visual models has fundamentally reshaped the perceptual quality landscape of AI-generated content~\cite{generate_aaai,iqa_aaai}.
Unlike traditional user-generated content (UGC) images, whose quality distributions are relatively stable~\cite{roy2023test}, modern generative models continuously shift the upper bound of perceptual quality within short time spans~\cite{t2i_bench}.
This evolution induces a perceptual drift: artifacts that were once salient become less common, while new failure modes emerge as models change.
This implies that evaluation must account for distribution shift and a drifting perceptual reference, rather than assuming a fixed perceptual scale~\cite{HINDER2023126640}.

Current image quality assessment (IQA) methods primarily rely on supervised learning with human mean opinion scores (MOS)~\cite{AGHIQA,MA_AGIQA}. While MOS can provide reliable subjective judgments under controlled protocols, it implicitly assumes a stable perceptual scale shared across data collection and deployment~\cite{AGIQA3K}.
This assumption breaks down in rapidly evolving generative settings, as illustrated in Figure~\ref{fig_1}, where identical MOS values may correspond to markedly different perceptual quality levels across model generations~\cite{HINDER2023126640}.
More fundamentally, maintaining valid MOS supervision would require frequent re-annotation to recalibrate the perceptual scale, resulting in limited long-term scalability and heavy annotation cost. 
As summarized in Table~\ref{tab_mos_cost}, this typically requires millions of human ratings even for medium-scale benchmarks.

These limitations have motivated exploration of unsupervised IQA methods, such as NSS-based metrics~\cite{BRISQUE,NIQE}, deep-feature statistics~\cite{MDFS}, reconstruction-based criteria~\cite{AdvCVAEIQA}, and CLIP-based measures~\cite{CLIP_IQA}. However, most existing approaches are designed for natural images or low-level distortions, and thus transfer poorly to modern AIGC~\cite{IPCE}. 
In practice, the quality of AI-generated content depends not only on low-level visual fidelity but also on whether the generated image correctly reflects the input prompt and avoids generation-specific artifacts~\cite{IPCE}. Such factors are difficult to capture with conventional unsupervised signals that mainly track low-level statistics. At the same time, although modern large multimodal models (MLLMs)~\cite{internvl3_5,Qwen3VL} demonstrate strong visual and semantic understanding, a key open challenge is how to derive scalable and continually updatable supervision that turns MLLMs into reliable label-free evaluators under perceptual drift.


\begin{table}[t]
\centering
\setlength{\tabcolsep}{6pt}
\caption{Human annotation scale of representative IQA benchmarks.}
\label{tab_mos_cost}
\resizebox{0.49\textwidth}{!}{
\begin{tabular}{lrr}
\toprule
Dataset & Images & Human Ratings \\
\midrule
KonIQ-10k \cite{KonIQ10k}           & 10,073    & 1,208,760 \\
PaQ-2-PiQ \cite{PaQ-2-PiQ}          & 40,000    & 4,000,000 \\
AIGIQA-20K \cite{AIGIQA20K}         & 20,000    & 420,000 \\
AGIN \cite{AGIN}                    & 6,049     & 907,350 \\
EvalMuse-40k \cite{evalmuse}        & 40,000    & 1,000,000 \\
EvalMi-50K \cite{LMM4LMM}           & 50,400    & 2,419,200 \\
Q-Eval-100k \cite{qeval100k}        & 100,000   & 960,000 \\
\bottomrule
\end{tabular}
}
\end{table}

In this work, we propose \textbf{ELIQ}, a \textit{\textbf{\underline{L}}abel-free Framework for \textbf{\underline{Q}}uality Assessment of \textbf{\underline{E}}volving AI-generated \textbf{\underline{I}}mages}.
Our key idea is to replace absolute MOS supervision with automatically constructed relative comparisons that can be periodically refreshed under perceptual drift.
ELIQ targets two model-agnostic dimensions of AIGC quality: visual quality and prompt-image alignment.
By generating high-quality positives and aspect-aware negative pairs that cover both conventional low-level distortions and AIGC-specific failure modes, ELIQ provides supervision independent of any fixed perceptual scale.
Based on this supervision, we fine-tune a pretrained multimodal model into a quality-aware critic, and then train a lightweight scoring module with gated visual-alignment representations and a Quality Query Transformer to predict visual and alignment quality scores.

Experiments on multiple AIGC and UGC benchmarks show that our method consistently outperforms existing unsupervised and label-free approaches and substantially narrows the gap with supervised baselines.
Moreover, ELIQ generalizes seamlessly from AIGC to UGC without architectural modification, providing a scalable alternative to MOS-dependent evaluation and enabling more sustainable quality assessment under evolving generative models. 

Our main contributions are threefold:
\begin{itemize}
\item We introduce \textbf{ELIQ}, a label-free framework for quality assessment of evolving AI-generated images, which decouples supervision from fixed MOS scales and remains effective as generative models and perceptual standards evolve.

\item We develop an assessment pipeline that jointly models visual quality and prompt-image alignment, using automatically constructed positive and aspect-specific negative pairs to adapt a pretrained multimodal model into a quality-aware critic with lightweight, label-free scoring.

\item  Extensive experiments across multiple benchmarks show that ELIQ consistently outperforms existing label-free and weakly supervised methods, remains competitive with strong supervised baselines, and generalizes effectively from AI-generated to user-generated image scenarios.
\end{itemize}

\section{Related Work}
\label{sec_related_works}

\subsection{AIGC Generation Models}
Recent text-to-image (T2I) models have rapidly evolved with diffusion and transformer-based backbones, continuously raising the perceptual quality ceiling and introducing diverse, model-specific artifacts.
Representative advances include DDPM~\cite{ddpm} and accelerated sampling with DDIM~\cite{ddim}, latent diffusion~\cite{LDMs} that underpins Stable Diffusion and SDXL~\cite{SDXL}, and transformer-style diffusion backbones such as DiT~\cite{DiT} and PixArt-$\alpha$~\cite{pixartalpha}.
Recent large-scale systems further scale data and unified conditioning, e.g., FLUX~\cite{FLUX} and Qwen-Image~\cite{qwenimage}.
This fast iteration makes it increasingly difficult to maintain consistent human annotations over time, motivating scalable supervision beyond MOS.

\subsection{Image Quality Assessment}
Early IQA methods estimate perceptual quality using handcrafted priors such as structural similarity (SSIM) and natural scene statistics (NSS), with representative no-reference metrics including BRISQUE~\cite{BRISQUE} and NIQE~\cite{NIQE}.
With deep learning, IQA shifts to learned quality representations that better handle authentic distortions; transformer-based models such as MUSIQ~\cite{MUSIQ} and hybrid designs like MANIQA~\cite{MANIQA} further improve multi-scale context modeling.

\subsection{Multi-modal Model-based IQA}
Vision-language priors enable IQA with stronger semantic awareness and improved zero-shot capability.
CLIP-based methods such as LIQE~\cite{LIQE} and CLIP-IQA~\cite{CLIP_IQA} explore embedding alignment and prompt-based assessment for no-reference quality prediction.
For AIGC-IQA, several approaches explicitly model prompt-image correspondence, e.g., IPCE~\cite{IPCE} and CLIP-AGIQA~\cite{CLIP_AGIQA}, while MLLM-based critics have been used to output discrete judgments or continuous scores for both quality and alignment~\cite{Zhang_CVPR}.
Despite their progress, most AIGC-IQA methods still rely on large-scale MOS or preference labels, which are costly to refresh under rapidly evolving generative models.

\subsection{Label-Free IQA Methods}
To reduce the dependence on MOS, label-free IQA estimates quality using proxies that avoid human scores.
Classical approaches follow an NSS paradigm, measuring deviations from learned pristine statistics, such as NIQE~\cite{NIQE} and IL-NIQE~\cite{ILNIQE}.
Recent deep variants learn distributional priors or ''distance-to-pristine'' criteria in feature/latent space~\cite{DSTS,AdvCVAEIQA,NROUQA}.
Another line learns quality-aware representations via self-supervision or pseudo-ranking, including QPT~\cite{QaPMBIQA}, CONTRIQUE~\cite{CONTRIQUE}, and ARNIQA~\cite{ARNIQA}.
Multimodal label-free IQA also emerges, where QualiCLIP~\cite{RWIQA} aligns synthetic degradations with quality-related text prompts to learn quality-sensitive embeddings without MOS.
However, existing label-free methods often rely on synthetic distortions or proxy supervision, limiting robustness to diverse and model-specific AIGC artifacts and motivating label-free supervision that explicitly constructs quality-aware comparisons.

\section{Proposed Method}

\subsection{Framework Overview}
\label{sec_overview}

\textbf{Problem setup.}
Given an AI-generated image $I$ and its prompt $p$, ELIQ aims to predict two quality scores:
a \emph{visual quality} score $\hat{s}_{\mathrm{vis}}(I)$ and an \emph{alignment quality} score $\hat{s}_{\mathrm{ali}}(I,p)$.
The key challenge is that MOS-based absolute supervision is expensive and quickly becomes outdated as generative models evolve.
ELIQ therefore replaces MOS with label-free relative supervision that can be re-generated for new model eras.

\textbf{Overall pipeline.}
ELIQ consists of two coupled components: (i) \emph{Label-Free Supervision Construction} and (ii) \emph{Model Training}.
First, we construct a set of high-quality positives and aspect-specific negatives to yield comparison tuples
\begin{equation}
\mathcal{T}=\big(I^{+},\,I^{-}_{\mathrm{tec}},\,I^{-}_{\mathrm{aes}},\,p,\,p^{-}_{\mathrm{ali}}\big),
\end{equation}
where $I^{-}_{\mathrm{tec}}$ and $I^{-}_{\mathrm{aes}}$ degrade visual quality, and $p^{-}_{\mathrm{ali}}$ induces prompt-image mismatch.
This construction covers both conventional corruptions and AIGC-specific failure modes.

Second, we use these tuples to adapt a pretrained multimodal model into a quality-aware critic and obtain quality scores.
Concretely, we (1) perform \emph{quality-aware instruction tuning} to obtain a backbone $\mathrm{MLLM}_{\theta^\ast}$ that can reason about technical quality, aesthetic quality, and alignment, and (2) freeze $\mathrm{MLLM}_{\theta^\ast}$ and train a lightweight scoring module to produce outputs with single-image inference:
\begin{equation}
(I,p)\ \longrightarrow\ \hat{s}_{\mathrm{vis}}(I),\ \hat{s}_{\mathrm{ali}}(I,p).
\end{equation}
The scoring module is trained only with pairwise ranking constraints derived from $\mathcal{T}$, encouraging $I^{+}$ to score higher than its visual negatives and $(I^{+},p)$ to score higher than $(I^{+},p^{-}_{\mathrm{ali}})$.
With this design, ELIQ remains independent of any fixed MOS scale while directly targeting the two key dimensions of AI-generated image quality.

\subsection{Label-Free Supervision Construction}
\label{sec_dataset}
\subsubsection{Prompt Selection}

To cover diverse visual content with clear semantics, we adopt a seven-category taxonomy: Indoor Scenes, Urban Scenes, Natural Scenes, People \& Activities, Objects \& Artifacts, Food, and Events.
Each category is expanded into representative sub-concepts, and we use GPT-5 with category-specific rules to generate $400$ text-to-image prompts that are diverse in wording yet consistent in semantics, such as realistic content, category-aligned constraints, and a length limit.
Since a prompt may involve multiple concepts, we further assign multi-label categories via a two-step procedure: rule-based keyword matching followed by an LLM refinement for uncertain cases.
More details on the taxonomy design, generation rules, and the multi-label assignment are provided in the Appendix.

\begin{figure*}[t]
    \centering
    \includegraphics[width=\textwidth]{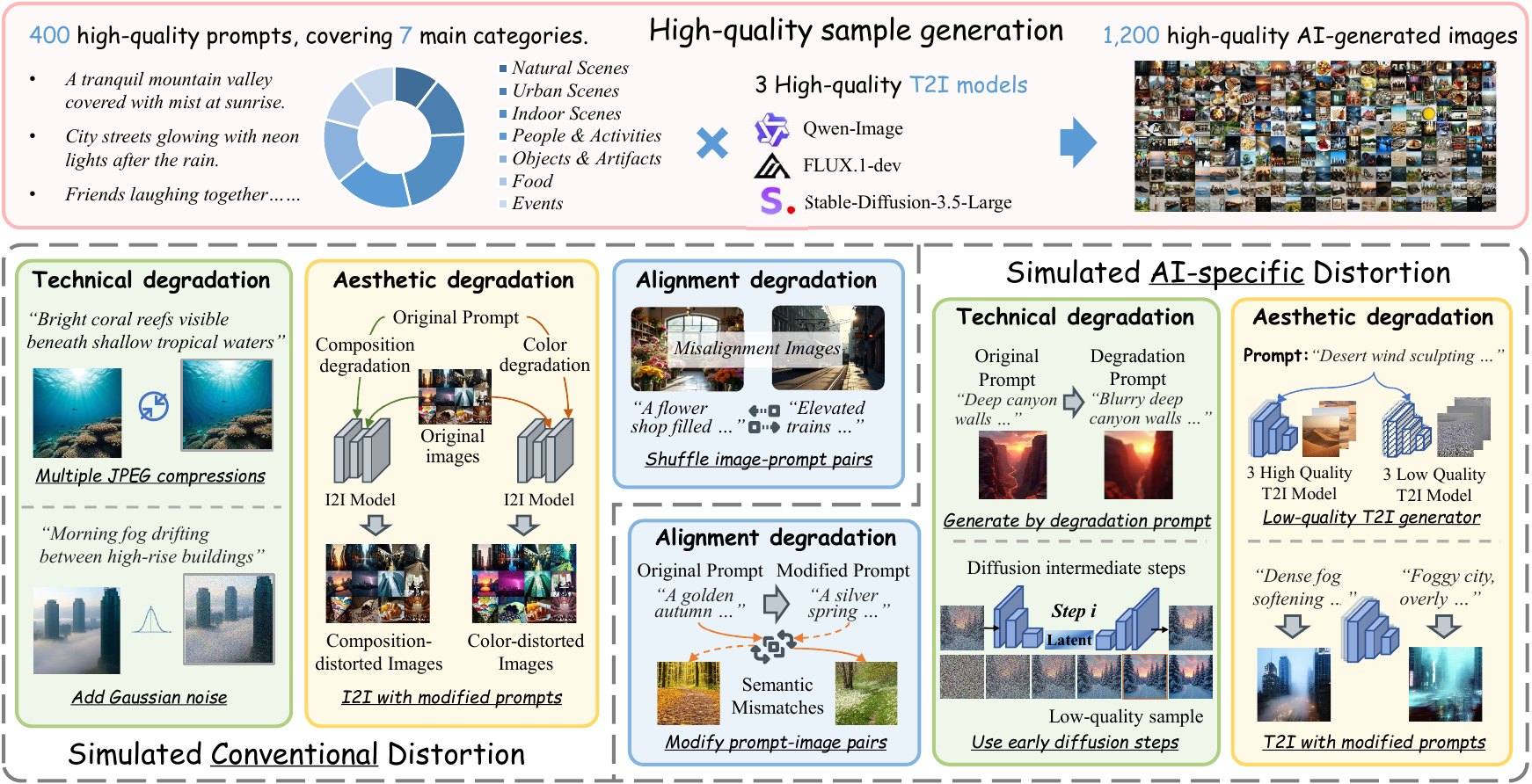}
    \caption{Overview of label-free positive and aspect-specific negative sample construction. High-quality images are generated from curated prompts using multiple T2I models, while negative samples are created by simulating technical, aesthetic, and alignment degradations, including both conventional distortions and AI-specific generation artifacts.
    }
    \label{fig_sample_construction}
\end{figure*}

\subsubsection{Positive Sample Generation}
To construct high-quality positive samples, we use the 400 curated prompts as inputs to three text-to-image generation models: Qwen-Image~\cite{qwenimage}, FLUX.1-dev~\cite{FLUX}, and Stable Diffusion 3.5-Large. 
These prompts cover seven major semantic categories, ensuring representative visual content.
These models differ in architecture and training data, allowing us to obtain image sets that reliably reflect the semantic content described in each prompt. 
For every prompt, one image is generated by each model, resulting in three corresponding sets of positive samples and a total of $1,200$ high-quality AI-generated images.

\subsubsection{Generation of Negative Samples}
\label{gen_neg}

To train our label-free IQA model, we construct diverse negative samples that cover both conventional image corruptions and AIGC-specific failure modes. 
We design two complementary degradation families, \textbf{conventional} and \textbf{AI-specific}, each instantiated along three dimensions: \textbf{technical}, \textbf{aesthetic}, and \textbf{alignment} quality.

For conventional degradations, technical negatives are synthesized via repeated JPEG compression and Gaussian noise. 
Aesthetic negatives are generated by applying controlled image-to-image editing, such as Qwen-Edit~\cite{qwen_image_edit} to high-quality images, inducing composition and color degradations while preserving the overall semantic content.
Alignment negatives are constructed by shuffling image-prompt pairs, creating semantic mismatches.

For AI-specific degradations, technical negatives are obtained by generating images with explicitly degraded prompts that induce low-fidelity structures, or by prematurely decoding intermediate diffusion latents.
Aesthetic negatives are produced via low-quality T2I generation, Stable Diffusion 1.1/1.4/1.5~\cite{sd_v1_1,sd_v1_4,sd_v1_5}, and via prompt modifications that lead to degraded visual appearance.
Alignment negatives are constructed by modified prompts to generate images with a high-quality T2I model, followed by mismatching images and prompts across prompt variants to induce semantic misalignment.
Detailed generation procedures and settings are provided in the Appendix.

\subsection{Model Training}
\label{sec_method}

\begin{figure*}[t]
    \centering
    \includegraphics[width=\textwidth]{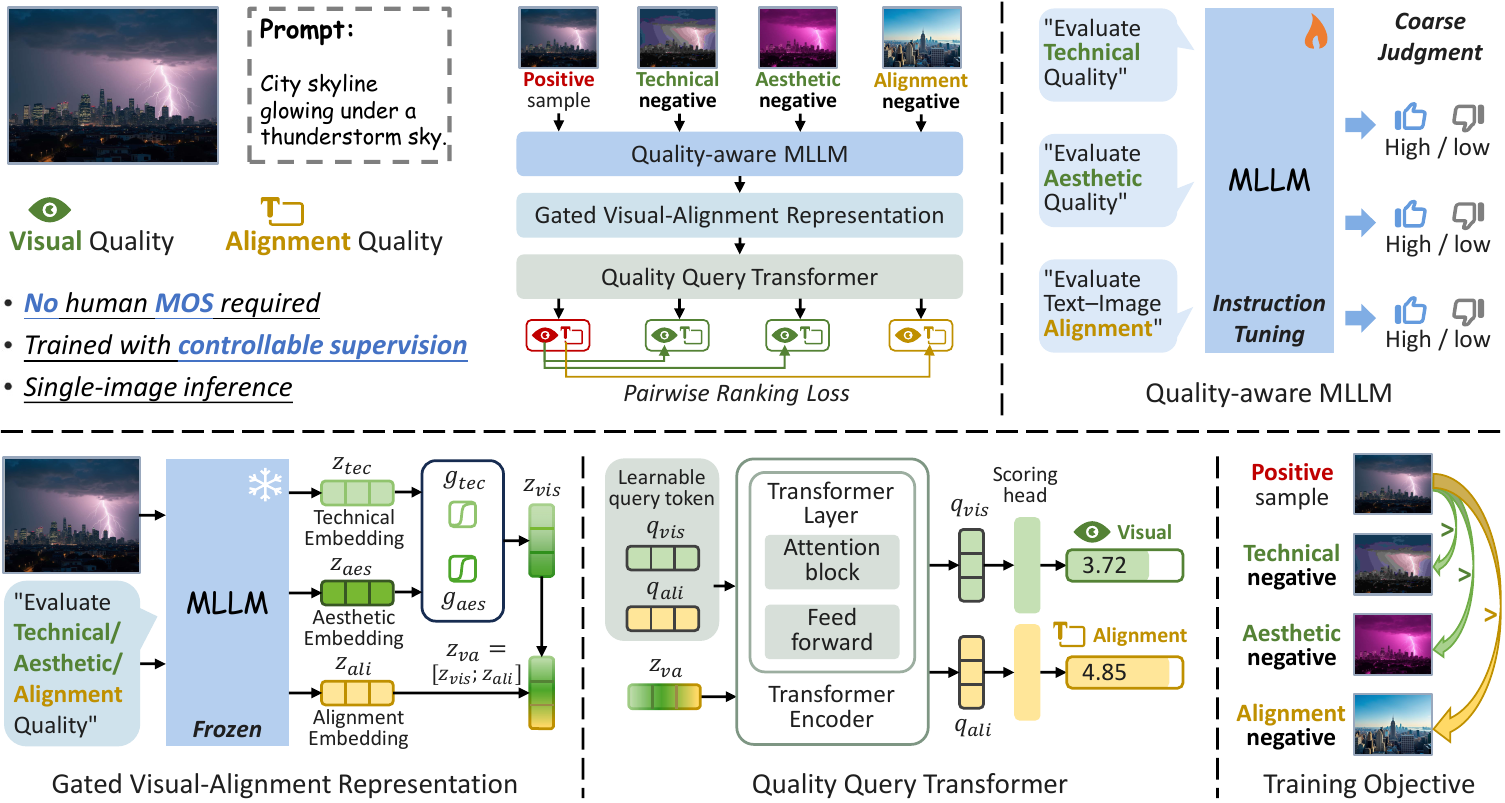}
    \caption{Overview of the proposed label-free visual and alignment quality scoring framework.}
    \label{fig_framework}
\end{figure*}

Our method contains two stages.
First, we perform \emph{Quality-aware Instruction Tuning} to adapt a pretrained MLLM into an aspect-aware backbone.
Second, we train a lightweight scoring module, termed \emph{Quality Query Transformer} (QQT), on top of the frozen aspect-aware embeddings. The schematic diagram of the proposed method is shown in Figure~\ref{fig_framework}.

\subsubsection{Quality-aware Instruction Tuning}
\label{subsec:instruction_tuning}

We adapt a pretrained MLLM into a quality-aware multimodal backbone that can explicitly reason about three perceptual aspects: technical quality, aesthetic quality, and text-image alignment.
We formulate each aspect-specific assessment as an instruction-following task and fine-tune the MLLM to output a discrete label $\texttt{low}$ or $\texttt{high}$.
No human MOS annotations are required; instead, we construct supervision using one positive image and aspect-specific negatives generated via controlled degradations.

\paragraph{Aspect-specific instructions and negatives.}
Given an AIGC image $I$ and its generation prompt $p$, where positive samples $I^{+}$ and aspect-specific negatives are constructed following Sec.~\ref{gen_neg}, we consider
\begin{equation}
d \in \{\mathrm{tec},\,\mathrm{aes},\,\mathrm{ali}\},
\end{equation}
corresponding to technical quality, aesthetic quality, and semantic alignment, respectively.
For each aspect $d$, we design an instruction $\mathrm{instr}_d$ that restricts the model to evaluate only that aspect while ignoring the others.
For \emph{technical} and \emph{aesthetic} quality, the instruction does not include the prompt to avoid injecting semantic cues.
For \emph{alignment}, the instruction includes the prompt $p$ as the reference text.

For each positive image $I^{+}$, we synthesize two visually degraded negatives
\begin{equation}
I^{-}_{\mathrm{tec}},\quad I^{-}_{\mathrm{aes}},
\end{equation}
where each negative is primarily degraded along its corresponding aspect.
For alignment, instead of modifying the image, we construct a mismatched prompt $p^{-}_{\mathrm{ali}}$ (e.g., sampled from another instance) and form an alignment-negative pair $(I^{+}, p^{-}_{\mathrm{ali}})$.

We assign pseudo-labels such that $I^{+}$ is labeled \texttt{high} for all three aspects, while
$I^{-}_{\mathrm{tec}}$ and $I^{-}_{\mathrm{aes}}$ are labeled \texttt{low} only for their primary aspects, and $(I^{+}, p^{-}_{\mathrm{ali}})$ is labeled \texttt{low} only for alignment.
We do not impose supervision on the non-primary aspects because they are not guaranteed by construction.

\paragraph{Instruction-tuning objective.}
We format each query in the standard MLLM chat style.
For aspect $d$, the input contains the embedded image and $\mathrm{instr}_d$; for $d=\mathrm{ali}$, the prompt is included as instructed, while for $d\in\{\mathrm{tec},\mathrm{aes}\}$ it is omitted.
The target response is a single label token $y_d\in\{\texttt{low},\texttt{high}\}$.
Let $\mathbf{x}_d$ and $\mathbf{y}_d$ denote the input and output token sequences.
We fine-tune the MLLM parameters $\theta$ by minimizing the autoregressive negative log-likelihood:
\begin{equation}
\mathcal{L}_{\mathrm{SFT}}
=
- \mathbb{E}_{(\cdot,d)}
\left[
\sum_{t=1}^{T_d}
\log p_{\theta}\Big(
y_{d,t}
\,\big|\,
\mathbf{x}_d,\,
\mathbf{y}_{d,<t}
\Big)
\right].
\end{equation}
After instruction tuning, we freeze the backbone parameters and use it as an aspect-aware feature extractor.

\subsubsection{Gated Visual-Alignment Representation}
\label{subsec:gated_rep}

Discrete $\texttt{low/high}$ outputs are too coarse for fine-grained quality prediction.
We therefore extract aspect-aware embeddings from the frozen backbone and build a compact representation tailored to two targets: visual quality and alignment quality.

\paragraph{Frozen aspect-aware embeddings.}
Let $\mathrm{MLLM}_{\theta^\ast}$ be the instruction-tuned backbone with frozen parameters $\theta^\ast$.
For each image $I$, we query the backbone with $\mathrm{instr}_d$ and extract the last-layer hidden state of the final token of input as a continuous embedding.
For technical and aesthetic aspects, we omit the prompt:
\begin{equation}
\mathbf{z}_{\mathrm{tec}}(I)=\mathrm{MLLM}_{\theta^\ast}\big(I,\,\mathrm{instr}_{\mathrm{tec}}\big)\in\mathbb{R}^{h},
\end{equation}
\begin{equation}
\mathbf{z}_{\mathrm{aes}}(I)=\mathrm{MLLM}_{\theta^\ast}\big(I,\,\mathrm{instr}_{\mathrm{aes}}\big)\in\mathbb{R}^{h},
\end{equation}
and for alignment, we condition on the prompt:
\begin{equation}
\mathbf{z}_{\mathrm{ali}}(I,p)=\mathrm{MLLM}_{\theta^\ast}\big(I,\,p,\,\mathrm{instr}_{\mathrm{ali}}\big)\in\mathbb{R}^{h},
\end{equation}
where $h$ is the backbone hidden size.
During training, we obtain alignment-negative embeddings using the mismatched prompt $p^{-}_{\mathrm{ali}}$, i.e., $\mathbf{z}_{\mathrm{ali}}(I^{+},p^{-}_{\mathrm{ali}})$.

\paragraph{Gated fusion for visual quality.}
Instead of predicting separate technical and aesthetic scores, we fuse their embeddings into a unified visual representation.
Given $\mathbf{z}_{\mathrm{tec}}(I)$ and $\mathbf{z}_{\mathrm{aes}}(I)$, we compute two element-wise gates:
\begin{equation}
\mathbf{g}_{\mathrm{tec}}=\sigma\!\left(\mathbf{W}_{\mathrm{tec}}[\mathbf{z}_{\mathrm{tec}};\mathbf{z}_{\mathrm{aes}}]\right),\qquad
\mathbf{g}_{\mathrm{aes}}=\sigma\!\left(\mathbf{W}_{\mathrm{aes}}[\mathbf{z}_{\mathrm{tec}};\mathbf{z}_{\mathrm{aes}}]\right),
\end{equation}
where $[\cdot;\cdot]$ denotes concatenation and $\sigma(\cdot)$ is the sigmoid function.
We modulate each branch and produce a fused embedding:
\begin{equation}
\tilde{\mathbf{z}}_{\mathrm{tec}}=\mathbf{z}_{\mathrm{tec}}\odot \mathbf{g}_{\mathrm{tec}},\qquad
\tilde{\mathbf{z}}_{\mathrm{aes}}=\mathbf{z}_{\mathrm{aes}}\odot \mathbf{g}_{\mathrm{aes}},
\end{equation}
\begin{equation}
\mathbf{z}_{\mathrm{vis}}(I)=\phi\big([\tilde{\mathbf{z}}_{\mathrm{tec}};\tilde{\mathbf{z}}_{\mathrm{aes}}]\big)\in\mathbb{R}^{h},
\end{equation}
where $\odot$ is element-wise multiplication and $\phi(\cdot)$ is a small MLP mapping $\mathbb{R}^{2h}\!\rightarrow\!\mathbb{R}^{h}$.

\subsubsection{Visual-alignment concatenation.}
We concatenate the fused visual embedding and the alignment embedding to form the final content token for scoring:
\begin{equation}
\mathbf{z}_{\mathrm{va}}(I,p)=[\mathbf{z}_{\mathrm{vis}}(I);\mathbf{z}_{\mathrm{ali}}(I,p)]\in\mathbb{R}^{2h}.
\end{equation}

\subsubsection{Quality Query Transformer}
\label{subsec:qqt}

We introduce a lightweight Transformer encoder on top of $\mathbf{z}_{\mathrm{va}}$, termed Quality Query Transformer (QQT), to predict visual and alignment quality scores.
QQT is trained with ranking supervision derived from label-free constructions, while keeping inference strictly single-image.
It adopts two learnable query tokens that attend to the content token and extract task-specific evidence for visual and alignment scoring.

Given an input pair $(I,p)$, we construct a 3-token sequence
\begin{equation}
\mathbf{X}_3(I,p)=
\big[
\mathbf{z}_{\mathrm{va}}(I,p),\;
\mathbf{q}_{\mathrm{vis}},\;
\mathbf{q}_{\mathrm{ali}}
\big]\in\mathbb{R}^{3\times 2h},
\end{equation}
where $\mathbf{q}_{\mathrm{vis}},\mathbf{q}_{\mathrm{ali}}\in\mathbb{R}^{2h}$ are learnable query tokens shared across all samples.
We project tokens and add learnable positional embeddings:
\begin{equation}
\mathbf{H}_3(I,p)=
\mathrm{TE}\!\left(
\mathbf{X}_3(I,p)\mathbf{W}_{\mathrm{proj}}^\top
+
\mathbf{E}_{\mathrm{pos}}
\right)\in\mathbb{R}^{3\times d_{\mathrm{model}}},
\end{equation}
where $\mathrm{TE}$ denotes the Transformer encoder.
We compute scores by applying two MLP heads, $f_{\mathrm{vis}}$ and $f_{\mathrm{ali}}$, to the corresponding query-token outputs:
\begin{equation}
\hat{s}_{\mathrm{vis}}(I)=f_{\mathrm{vis}}\big(\mathbf{h}^{\,q}_{\mathrm{vis}}(I,p)\big),\qquad
\hat{s}_{\mathrm{ali}}(I,p)=f_{\mathrm{ali}}\big(\mathbf{h}^{\,q}_{\mathrm{ali}}(I,p)\big),
\end{equation}
where $\mathbf{h}^{\,q}_{\mathrm{vis}}(I,p)$ and $\mathbf{h}^{\,q}_{\mathrm{ali}}(I,p)$ denote the encoded outputs of the visual-query token and alignment-query token in $\mathbf{H}_3(I,p)$, respectively.

\subsubsection{Training Objective}
\label{subsec:training_objective}

We train the gated fusion module and QQT using margin-based ranking losses derived from label-free construction.
For each positive pair $(I^{+},p)$, we construct two visually degraded negatives $I^{-}_{\mathrm{tec}}$ and $I^{-}_{\mathrm{aes}}$, and one mismatched prompt $p^{-}_{\mathrm{ali}}$ for alignment:
\begin{equation}
\mathcal{T}=\big(I^{+},\,I^{-}_{\mathrm{tec}},\,I^{-}_{\mathrm{aes}},\,p,\,p^{-}_{\mathrm{ali}}\big).
\end{equation}
All scores are predicted independently by QQT:
\begin{equation}
\hat{s}_{\mathrm{vis}}(I^{+}),\;
\hat{s}_{\mathrm{vis}}(I^{-}_{\mathrm{tec}}),\;
\hat{s}_{\mathrm{vis}}(I^{-}_{\mathrm{aes}}),\;
\hat{s}_{\mathrm{ali}}(I^{+},p),\;
\hat{s}_{\mathrm{ali}}(I^{+},p^{-}_{\mathrm{ali}}).
\end{equation}

\paragraph{Visual ranking loss.}
Since both $I^{-}_{\mathrm{tec}}$ and $I^{-}_{\mathrm{aes}}$ are visually degraded, we supervise visual quality with two pairwise constraints (instead of hard-negative aggregation) to avoid gradient sparsity:
\begin{equation}
\begin{aligned}
\mathcal{L}_{\mathrm{vis}}
=
\mathbb{E}_{\mathcal{T}}\Big[
&\max\big(0,\, m - \hat{s}_{\mathrm{vis}}(I^{+}) + \hat{s}_{\mathrm{vis}}(I^{-}_{\mathrm{tec}})\big) \\
&+ \max\big(0,\, m - \hat{s}_{\mathrm{vis}}(I^{+}) + \hat{s}_{\mathrm{vis}}(I^{-}_{\mathrm{aes}})\big)
\Big].
\end{aligned}
\end{equation}

\paragraph{Alignment ranking loss.}
For alignment, we enforce that the positive image aligns better with its original prompt than with a mismatched prompt:
\begin{equation}
\mathcal{L}_{\mathrm{ali}}
=
\mathbb{E}_{\mathcal{T}}\Big[
\max\big(0,\, m - \hat{s}_{\mathrm{ali}}(I^{+},p) + \hat{s}_{\mathrm{ali}}(I^{+},p^{-}_{\mathrm{ali}})\big)
\Big].
\end{equation}

\paragraph{Overall objective.}
The final loss is
\begin{equation}
\mathcal{L}
=
\mathcal{L}_{\mathrm{vis}}
+
\mathcal{L}_{\mathrm{ali}}.
\end{equation}



\begin{table*}[t]
\caption{Performance comparison of ELIQ with supervised, weak-supervised, and label-free methods on three AIGC benchmarks and two UGC benchmarks. The best-performing metric is highlighted in bold, and the second-best is underlined within each supervision setting.}
\label{tab_performance_visual}
\centering
\resizebox{\textwidth}{!}{
\begin{tabular}{l c cc cc cc !{\vrule width 0.6pt} cc cc}
\toprule
\multirow{3}{*}{Methods} & \multirow{3}{*}{Type}
& \multicolumn{6}{c}{\textbf{AIGC Datasets}}
& \multicolumn{4}{c}{\textbf{UGC Datasets}} \\
\cmidrule(lr){3-8}\cmidrule(lr){9-12}
& 
& \multicolumn{2}{c}{AGIQA-3K}
& \multicolumn{2}{c}{AIGCIQA2023}
& \multicolumn{2}{c}{AIGIQA-20K}
& \multicolumn{2}{c}{KonIQ-10k}
& \multicolumn{2}{c}{SPAQ} \\
\cmidrule(lr){3-4}\cmidrule(lr){5-6}\cmidrule(lr){7-8}\cmidrule(lr){9-10}\cmidrule(lr){11-12}
& 
& \multicolumn{2}{c}{SRCC\quad PLCC}
& \multicolumn{2}{c}{SRCC\quad PLCC}
& \multicolumn{2}{c}{SRCC\quad PLCC}
& \multicolumn{2}{c}{SRCC\quad PLCC}
& \multicolumn{2}{c}{SRCC\quad PLCC} \\
\midrule
\multicolumn{12}{c}{\textbf{\textit{Supervised}}} \\
BRISQUE ~\cite{BRISQUE}     & handcraft   & 0.472 & 0.561 & 0.446 & 0.465 & 0.466 & 0.558 & 0.705 & 0.707 & 0.802 & 0.805 \\
HyperIQA ~\cite{HyperIQA}  & data-driven & 0.850 & 0.904 & 0.822 & 0.852 & 0.816 & 0.832 & 0.904 & 0.915 & 0.915 & 0.918 \\
MANIQA ~\cite{MANIQA}      & data-driven & 0.861 & 0.911 & 0.818 & 0.847 & 0.850 & 0.887 & 0.930 & 0.946 & 0.922 & 0.927 \\
DBCNN ~\cite{DBCNN}       & data-driven & 0.826 & 0.890 & 0.807 & 0.829 & 0.805 & 0.848 & 0.844 & 0.862 & 0.909 & 0.927 \\
MUSIQ ~\cite{MUSIQ}        & data-driven & 0.820 & 0.865 & 0.803 & 0.820 & 0.832 & 0.864 & 0.824 & 0.937 & 0.873 & 0.868 \\
StairIQA ~\cite{StairIQA} & data-driven & 0.834 & 0.893 & 0.808 & 0.832 & 0.789 & 0.842 & 0.920 & 0.936 & 0.923 & 0.929 \\
Q-Align ~\cite{qalign}     & data-driven & 0.852 & 0.881 & 0.841 & 0.860 & 0.874 & 0.889 & 0.922 & 0.911 & 0.887 & 0.886 \\
MA-AGIQA ~\cite{MA_AGIQA} & data-driven & 0.893 & 0.927 & 0.853 & 0.856 & 0.864 & 0.905 & 0.933 & 0.948 & 0.927 & 0.932 \\
\midrule

\multicolumn{12}{c}{\textbf{\textit{Weak-supervised}}} \\
CONTRIQUE ~\cite{CONTRIQUE} & data-driven & 0.817 & 0.879 & 0.771 & 0.789 & 0.788 & 0.807 & 0.894 & 0.906 & 0.914 & 0.919 \\
Re-IQA ~\cite{REIQA}       & data-driven & 0.811 & 0.874 & 0.769 & 0.789 & 0.787 & 0.811 & \textbf{0.914} & \underline{0.923} & \textbf{0.918} & \textbf{0.925} \\
CLIP-IQA+ ~\cite{CLIP_IQA} & data-driven & \underline{0.844} & \underline{0.894} & \underline{0.817} & \underline{0.835} & \underline{0.833} & \underline{0.854} & 0.895 & 0.909 & 0.864 & 0.866 \\
ARNIQA ~\cite{ARNIQA}      & data-driven & 0.803 & 0.881 & 0.754 & 0.764 & 0.778 & 0.792 & 0.869 & 0.883 & 0.904 & 0.909 \\
GRepQ-D ~\cite{GRepQ}      & data-driven & 0.807 & 0.858 & 0.767 & 0.783 & 0.789 & 0.810 & 0.855 & 0.868 & 0.903 & 0.917 \\
\textbf{Ours}                    & data-driven & \textbf{0.876} & \textbf{0.911} & \textbf{0.837} & \textbf{0.851} & \textbf{0.856} & \textbf{0.883} & \underline{0.912} & \textbf{0.924} & \underline{0.915} & \textbf{0.925} \\
\midrule

\multicolumn{12}{c}{\textbf{\textit{Label-free}}} \\
NIQE ~\cite{NIQE}          & handcraft   & 0.523 & 0.566 & 0.511 & 0.523 & 0.208 & 0.337 & 0.551 & 0.488 & 0.703 & 0.670 \\
ILNIQE ~\cite{ILNIQE}       & handcraft   & 0.609 & 0.655 & 0.594 & 0.611 & 0.335 & 0.455 & 0.453 & 0.467 & 0.719 & 0.654 \\
CLIP-IQA ~\cite{CLIP_IQA}  & data-driven & 0.638 & 0.711 & 0.589 & 0.604 & 0.388 & 0.537 & 0.695 & 0.727 & 0.738 & 0.735 \\
MDFS ~\cite{MDFS}           & data-driven & 0.672 & 0.676 & 0.659 & 0.667 & 0.691 & 0.695 & 0.733 & 0.737 & 0.741 & 0.754 \\
QualiCLIP ~\cite{QualiCLIP} & data-driven & 0.667 & \underline{0.735} & 0.646 & 0.661 & 0.679 & 0.694 & \underline{0.817} & \textbf{0.838} & \underline{0.841} & \textbf{0.851} \\
GRepQ-Z ~\cite{GRepQ}      & data-driven & 0.613 & 0.734 & 0.602 & 0.612 & 0.624 & 0.634 & 0.768 & 0.784 & 0.823 & 0.839 \\
DUBMA ~\cite{DUBMA}       & data-driven & \underline{0.684} & 0.701 & \underline{0.671} & \underline{0.680} & \underline{0.695} & \underline{0.697} & 0.703 & 0.740 & 0.834 & \underline{0.841} \\
\textbf{Ours}                    & data-driven & \textbf{0.801} & \textbf{0.827} & \textbf{0.767} & \textbf{0.781} & \textbf{0.786} & \textbf{0.803} & \textbf{0.818} & \underline{0.831} & \textbf{0.842} & \textbf{0.851} \\
\bottomrule
\end{tabular}
}
\end{table*}



\subsubsection{Inference}
\label{subsec:inference}

Given $(I,p)$, we extract frozen embeddings $\mathbf{z}_{\mathrm{tec}}(I)$, $\mathbf{z}_{\mathrm{aes}}(I)$, and $\mathbf{z}_{\mathrm{ali}}(I,p)$, fuse technical and aesthetic embeddings into $\mathbf{z}_{\mathrm{vis}}(I)$, and form $\mathbf{z}_{\mathrm{va}}(I,p)$.
We then construct $\mathbf{X}_3(I,p)=[\mathbf{z}_{\mathrm{va}}(I,p), \mathbf{q}_{\mathrm{vis}}, \mathbf{q}_{\mathrm{ali}}]$ and predict two scores:
\begin{equation}
\hat{s}_{\mathrm{vis}}(I),\quad \hat{s}_{\mathrm{ali}}(I,p).
\end{equation}
Optionally, when a small labeled validation split is available, we apply a monotonic linear calibration to map raw scores to the MOS range. This post-hoc scaling does not affect rank-based evaluation.

\section{Experiments}

\subsection{Datasets and Experimental Settings}
\label{subsec:exp_setting}

We evaluate our method on three AIGC benchmarks, AGIQA-3K~\cite{AGIQA3K}, AIGCIQA2023~\cite{AIGCIQA2023}, and AIGIQA-20K~\cite{AIGIQA20K}, as well as two UGC IQA benchmarks, KonIQ-10k~\cite{KonIQ10k} and SPAQ~\cite{spaq}. Details of dataset statistics and splits are provided in the Appendix.

We report performance in two settings.
\emph{(i) Label-free.} We directly use the trained model to produce visual and alignment scores without any MOS supervision, and report the correlation with human ratings.
\emph{(ii) Weak-supervised.} To measure how well our learned representations transfer to human opinion scores, we append a lightweight linear regressor on the QQT and fine-tune only this linear layer using a small portion of MOS-labeled data (20\% for AIGC benchmarks and 30\% for UGC benchmarks), while keeping all other modules frozen and evaluating on the remaining dataset.

All experiments are conducted using Qwen3-VL-8B-Instruct as the MLLM backbone. 
We evaluate the consistency between predicted quality scores $\{\hat{y}_i\}_{i=1}^N$ and ground-truth scores $\{y_i\}_{i=1}^N$ using two correlation coefficients: Spearman’s $\rho$ (SRCC) and Pearson’s $r$ (PLCC). 
More implementation details are provided in the Appendix.

\subsection{Performance Analysis}

The proposed models are trained on the dataset introduced in Section \ref{sec_dataset}. 
The results are summarized in Table~\ref{tab_performance_visual} for visual quality on both AIGC and UGC benchmarks, and in Table~\ref{tab_performance_alignment} for alignment quality on AIGC benchmarks.




\subsubsection{Visual Quality}

In the weak-supervised setting, our method achieves the best performance on all three AIGC benchmarks, with SRCC of $0.876$ (AGIQA-3K), $0.837$ (AIGCIQA2023), and $0.856$ (AIGIQA-20K), remaining competitive with fully supervised approaches.
In the \emph{label-free} setting, it consistently outperforms existing label-free and handcrafted baselines, reaching SRCC of $0.801$, $0.767$, and $0.786$ on the same datasets without any target-domain MOS fine-tuning.

On UGC benchmarks, our method attains SRCC of $0.912$ on KonIQ-10k and $0.915$ on SPAQ using only $30\%$ MOS-labeled data. In contrast, the strongest baseline Re-IQA requires $70\%$ for slightly higher correlation.
Without any MOS labels, our label-free variant still achieves SRCC of $0.818$ (KonIQ-10k) and $0.842$ (SPAQ), demonstrating robust and scalable quality prediction and effective transfer from AIGC to real-world User-Generated Content.

\subsubsection{Alignment Quality}

Table~\ref{tab_performance_alignment} presents prompt-image alignment performance on AGIQA-3K and AIGCIQA2023.
In the weak-supervised setting, our method achieves the highest correlation on both datasets, while in the label-free setting, it still reaches SRCC of $0.717$ on AGIQA-3K and $0.712$ on AIGCIQA2023 without using any human annotations, consistently outperforming all label-free baselines.

\begin{table}
\caption{The performance of the proposed ELIQ method and the compared supervised and label-free alignment quality. The best-performing metric is highlighted in bold.}
\label{tab_performance_alignment}
\resizebox{0.48\textwidth}{!}{
\begin{tabular}{lllll}
\toprule
\multirow{2}{*}{Methods}  & \multicolumn{2}{c}{AGIQA-3k} & \multicolumn{2}{c}{AIGCIQA2023} \\ 
\cmidrule(lr){2-3} \cmidrule(lr){4-5}
& SRCC & PLCC & SRCC & PLCC \\
\midrule
\multicolumn{5}{c}{\textbf{\textit{Weak-supervised}}}                                                         \\
CLIP (AAAI, 2023)           & 0.597            & 0.683            & 0.617              & 0.623           \\
CLIP-IQA+ (AAAI, 2023)      & 0.704            & 0.738            & 0.729              & 0.736           \\
ImageReward (NIPS, 2023)    & 0.729            & 0.786            & 0.749              & 0.759           \\
HPSv1 (ICCV, 2023)          & 0.634            & 0.700            & 0.663              & 0.674           \\
PickScore (NIPS, 2023)      & 0.697            & 0.763            & 0.716              & 0.734           \\
StairReward (TCSVT, 2023)   & 0.747            & 0.852            & 0.760              & 0.777           \\
IPCE (CVPR, 2024)           & 0.770            & 0.872            & 0.797              & 0.788           \\
\textbf{Ours}               & \textbf{0.789}   & \textbf{0.881}   & \textbf{0.804}     & \textbf{0.811}  \\
\midrule                                                                                                 
\multicolumn{5}{c}{\textbf{\textit{Label-free}}}                                                         \\
CLIP (AAAI, 2023)           & 0.428            & 0.450            & 0.463              & 0.474           \\
CLIP-IQA+ (AAAI, 2023)      & 0.501            & 0.524            & 0.521              & 0.538           \\
ImageReward (NIPS, 2023)    & 0.579            & 0.607            & 0.589              & 0.609           \\
HPSv1 (ICCV, 2023)          & 0.562            & 0.577            & 0.597              & 0.626           \\
PickScore (NIPS, 2023)      & 0.593            & 0.622            & 0.626              & 0.652           \\
\textbf{Ours}               & \textbf{0.717}    & \textbf{0.730}  & \textbf{0.712}     & \textbf{0.711}  \\
\bottomrule
\end{tabular}
}
\centering
\end{table}

\subsection{Ablation Study}
\label{sec:ablation}

We conduct a set of ablations to validate key components of ELIQ, reporting SRCC and PLCC for both visual quality and alignment quality.
We first validated the effectiveness of the degradation of negative samples.

\begin{table}
\caption{Ablation study of the negative degradation we used in ELIQ, where Conv and AI-spec denote Conventional degradation and AI-specific degradation, respectively.}
\label{tab_ablation_degradation_1}
\resizebox{0.48\textwidth}{!}{
\begin{tabular}{l|cc|cccc}
\toprule
\multirow{2}{*}{No.} & \multicolumn{2}{c}{Module} & \multicolumn{2}{c}{Visual} & \multicolumn{2}{c}{Alignment} \\ 
\cmidrule(lr){2-3}         \cmidrule(lr){4-5}                  \cmidrule(lr){6-7}                
 & Conv        & AI-spec   & SRCC            & PLCC           & SRCC           & PLCC           \\
\midrule                                                                                        
1 &  \ding{52} &           & 0.796           & 0.819          & 0.713          & 0.725          \\
2 &            & \ding{52} & 0.782           & 0.805          & 0.702          & 0.718          \\
3 &  \ding{52} & \ding{52} & \textbf{0.801}  & \textbf{0.827} & \textbf{0.717} & \textbf{0.730} \\
\bottomrule
\end{tabular}
}
\centering
\end{table}

TABLE~\ref{tab_ablation_degradation_1} shows that conventional negatives are more effective than AI-specific ones when used alone, indicating conventional distortion priors provide stable and transferable supervision. Using both together yields the best SRCC and PLCC on visual quality and alignment, confirming their complementarity.

TABLE~\ref{tab_ablation_degradation_2} further reveals that a single aspect is insufficient: \emph{Tec}-only/\emph{Aes}-only weakens visual prediction, while \emph{Ali}-only improves alignment but hurts visual quality. Combining two aspects helps, and including \emph{Ali} consistently benefits alignment; the full \emph{Tec+Aes+Ali} setting performs best overall.

In addition, we include four supplementary ablation studies to validate key design choices of ELIQ, as shown in Appendix. Specifically, they examine (i) the necessity of quality-aware instruction tuning for adapting a pretrained MLLM to quality assessment, (ii) the effectiveness of the proposed visual representation that combines technical and aesthetic cues, (iii) the role of the Quality Query Transformer in enabling task-specific quality prediction from a single image, and (iv) the impact of the proposed label-free training objective with multiple ranking constraints.


\begin{table}
\caption{Ablation study of the negative degradation we used in ELIQ, where Tec, Aes, and Ali denote technical degradation, aesthetic degradation, and alignment degradation, respectively.}
\label{tab_ablation_degradation_2}
\resizebox{0.48\textwidth}{!}{
\begin{tabular}{l|ccc|cccc}
\toprule
\multirow{2}{*}{No.} & \multicolumn{3}{c}{Module} & \multicolumn{2}{c}{Visual} & \multicolumn{2}{c}{Alignment}    \\ 
\cmidrule(lr){2-4}                      \cmidrule(lr){5-6}                 \cmidrule(lr){7-8}                       
& Tec         & Aes       & Ali       & SRCC            & PLCC           & SRCC           & PLCC                  \\
\midrule                                                                                                           
1 & \ding{52} &           &           & 0.773           & 0.794          & 0.703          & 0.712                 \\
2 &           & \ding{52} &           & 0.769           & 0.786          & 0.700          & 0.711                 \\
3 &           &           & \ding{52} & 0.751           & 0.772          & 0.712          & 0.723                 \\
4 & \ding{52} & \ding{52} &           & 0.789           & 0.807          & 0.710          & 0.719                 \\
5 &           & \ding{52} & \ding{52} & 0.787           & 0.804          & 0.715          & 0.727                 \\
6 & \ding{52} &           & \ding{52} & 0.796           & 0.818          & 0.714          & 0.728                 \\
7 & \ding{52} & \ding{52} & \ding{52} & \textbf{0.801}  & \textbf{0.827} & \textbf{0.717} & \textbf{0.730}        \\
\bottomrule
\end{tabular}
}
\centering
\end{table}


\section{Conclusion}
\label{sec_conclusion}

Overall, our findings indicate that MLLM-derived quality priors offer a practical and scalable alternative to traditional MOS-based supervision for AIGC quality assessment. Rather than treating quality evaluation as a static, human-anchored problem, our framework reframes it as a model-driven process that can continuously adapt to evolving generative capabilities.
This shift opens the door to sustainable quality assessment pipelines that keep pace with rapid advances in generative models, without incurring repeated annotation costs.

\section*{Impact Statement}
This paper presents work whose goal is to advance the field of Machine
Learning. There are many potential societal consequences of our work, none
which we feel must be specifically highlighted here.


\bibliography{example_paper}
\bibliographystyle{icml2026}

\end{document}